\documentclass[10pt, a4paper]{article}

\usepackage{lrec-coling2024}\usepackage{bbm}

\usepackage{url}
\usepackage{subfigure}
\usepackage{multirow}
\usepackage{enumitem}
\usepackage{stmaryrd}
\usepackage{array}
\usepackage{threeparttable}
\usepackage{blindtext}
\usepackage{amssymb}

\usepackage[ruled,vlined,linesnumbered,longend]{algorithm2e}
\usepackage{svg}
\usepackage{placeins}

\usepackage{soul}
\usepackage[utf8]{inputenc}
\usepackage{graphicx}
\usepackage{amsmath}
\usepackage{booktabs}
\usepackage{lipsum}
\usepackage{listings}
\usepackage{xcolor, soul}
\sethlcolor{blue!10}
\usepackage{colortbl}
\newcommand{\ourmetric}{\textsc{SQC-Score}\xspace}

\definecolor{sgreen}{HTML}{F3FADF} 
\definecolor{mgreen}{HTML}{E0EAB5} 
\definecolor{dgreen}{HTML}{CDDC8C} 
\definecolor{ddgreen}{HTML}{B8CF61} 

\definecolor{'sred'}{HTML}{FFEAE8}

\usepackage{amsmath,amsfonts,bm}

\def\eqref#1{equation~\ref{#1}}

\def\1{\bm{1}}

\DeclareMathAlphabet{\mathsfit}{\encodingdefault}{\sfdefault}{m}{sl}
\SetMathAlphabet{\mathsfit}{bold}{\encodingdefault}{\sfdefault}{bx}{n}

\title{
Evaluating Generative Language Models in \\
Information Extraction as Subjective Question Correction
}

\name{
    Yuchen Fan\textsuperscript{1}$^\heartsuit$,
    Yantao Liu\textsuperscript{2}$^\heartsuit$, 
    Zijun Yao\textsuperscript{3,4} \\
    \large\bf
    Jifan Yu\textsuperscript{5}, 
    Lei Hou\textsuperscript{3,4},
    and Juanzi Li\textsuperscript{3,4}$^\spadesuit$
    \thanks{
        $^\heartsuit$ Equal contribution.
        Work was done when they are interns at Tsinghua University.
    }
    \thanks{
        $^\spadesuit$ Corresponding author.
    }
}

\address{
\textsuperscript{1}Beijing University of Posts and Telecommunications, Beijing, China\\
\textsuperscript{2}University of Chinese Academy of Science, Beijing, China \\ 
$^3$BNRist; $^4$KIRC, Institute for Artificial Intelligence; $^5$Institute of Education \\
Tsinghua University, Beijing 100084, China \\
yaozj20@mails.tsinghua.edu.cn, \{houlei, lijuanzi\}@mails.tsinghua.edu.cn\\
}

\abstract{
Modern Large Language Models (LLMs) have showcased remarkable prowess in various tasks necessitating sophisticated cognitive behaviors. Nevertheless, a paradoxical performance discrepancy is observed, where these models underperform in seemingly elementary tasks like relation extraction and event extraction due to two issues in conventional evaluation.
(1) The imprecision of existing evaluation metrics that struggle to effectively gauge semantic consistency between model outputs and ground truth, and (2) The inherent incompleteness of evaluation benchmarks, primarily due to restrictive human annotation schemas, resulting in underestimated LLM performances.
Inspired by the principles in subjective question correction, we propose a new evaluation method, \ourmetric.
This method innovatively utilizes LLMs, fine-tuned through subjective question correction data, to refine matching between model outputs and golden labels. 
Additionally, by incorporating a Natural Language Inference (NLI) model, \ourmetric enriches golden labels, addressing benchmark incompleteness by acknowledging correct yet previously omitted answers. 
Results on three information extraction tasks show that \ourmetric is more preferred by human annotators than the baseline metrics.
Utilizing \ourmetric, we conduct a comprehensive evaluation of the state-of-the-art LLMs and provide insights for future research for information extraction. Dataset and associated codes can be accessed at our \href{https://github.com/THU-KEG/SQC-Score}{GitHub repository}.
\\ \newline 
}

\begin{document}

\maketitleabstract

\section{Introduction}

Modern large language models (LLMs)~\cite{gpt3,pythia,gpt4,touvron2023llama2} have demonstrated extraordinary capabilities in tasks even requiring high-order cognition behavior~\cite{bubeck2023sparks}.
However, they are reported~\cite{wei2023zero,gao2023exploring} to fall short on some elementary tasks, such as relation extraction and event extraction, thus forming a counter-intuitive performance discrepancy.

\begin{figure}
    \centering
    \includegraphics[width=0.9\linewidth]{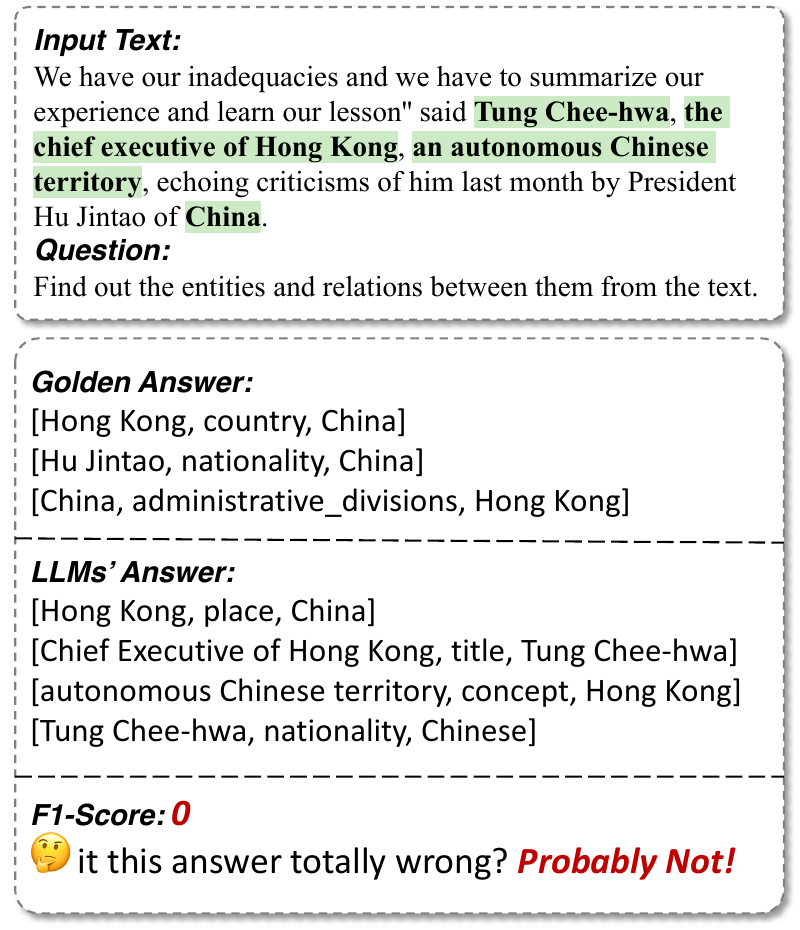}
       \caption{
    Exemplifying performance discrepancy in relation extraction task.
    (1) The triple \texttt{(Hong Kong,} \texttt{Country,} \texttt{China)} in golden answer shares the same semantics with the triple \texttt{(Hong Kong,} \texttt{Place,} \texttt{China)} output by LLM, but the conventional metrics would not treat them as the same.
    (2) The triple \texttt{(Tung Chee-hwa,} \texttt{nationality,} \texttt{Chinese)} in LLMs' output is plausible but not included in the golden label.
    }
    \label{fig:intro}
\end{figure}

We attribute this performance discrepancy to conservative task evaluation methods with two main factors:
\textbf{(1) Imprecision of Evaluation Metrics.} 
Modern LLMs flexibly generate natural language utterances as answers, while similarity-based metrics fail to evaluate semantic level consistency between model output and ground-truth answers.
\textbf{(2) Incompleteness of Evaluation Benchmarks.}
The construction of evaluation benchmarks highly relies on human annotation.
As a trade-off for annotation cost and precision, annotators are usually required to annotate answers based on a pre-defined schema, making relatively low golden label recall, which underestimates LLMs if they output answers beyond annotation.
Figure~\ref{fig:intro} exemplifies these problems.
Therefore, there is an emergent need for an advanced evaluation method.

\begin{figure*}[t]
    \hspace{-0.3cm}
       \includegraphics[width=\linewidth]{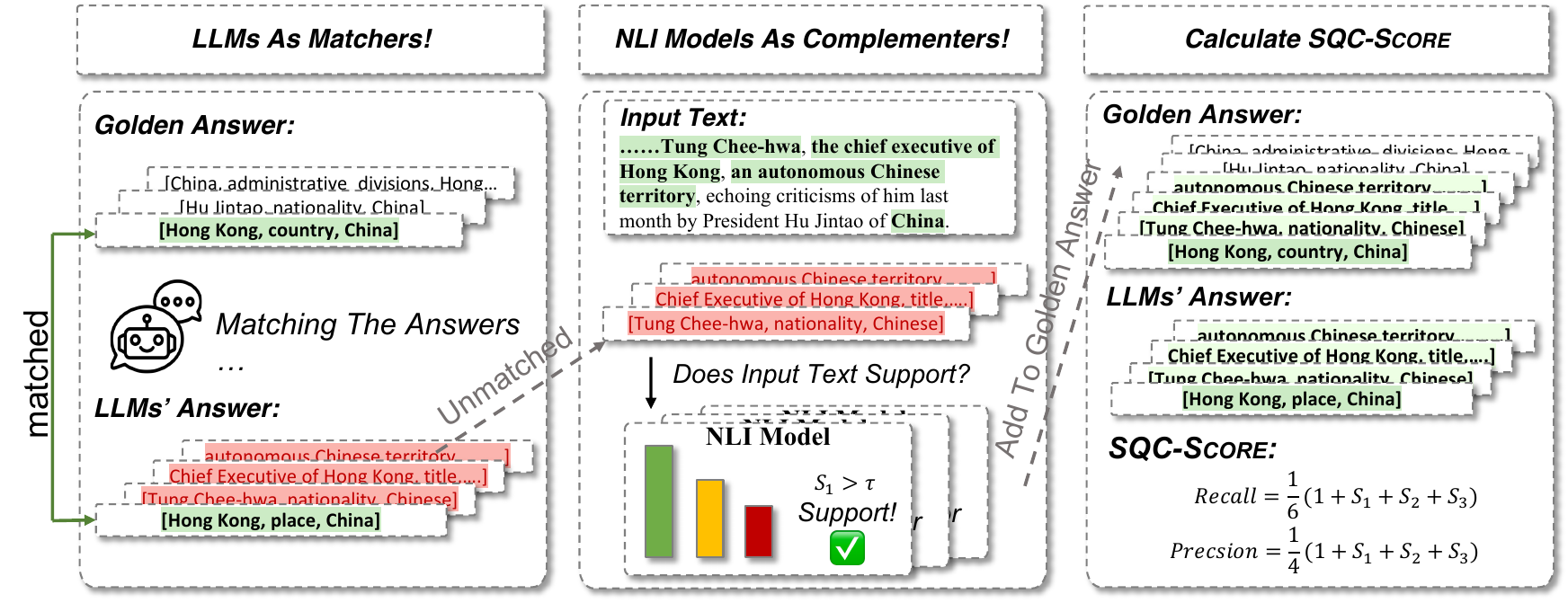}
    \vspace{-0.09in}
    \caption{The overall pipeline of \ourmetric,
    where $s_1, s_2, s_3$ are the entailment scores of the three unmatched triples and $\tau$ is the threshold for entailment.
    }
    \label{fig:pipeline}
\end{figure*}

There are efforts made to design new automatic evaluation methods.
For example, BERTScore \cite{bertscore} and BARTScore \cite{bartscore} leverage the contextual embeddings to evaluate the similarity between model output and golden label.
GPTScore~\cite{gptscore} evaluates the quality of generated text by comparing the perplexity of the generated text and the golden label.
However, they are mainly designed for text generation tasks, which focus more on fluency and consistency than correctness.
They also neglect the incompleteness of datasets.

Inspired by the way of \textbf{s}ubjective \textbf{q}uestion \textbf{c}orrection, which gives credit to answers with acceptable rationality, we propose \ourmetric.
For providing a precise evaluation score, we leverage LLMs supervised-fine-tuned on subjective question correction data to match between model output and golden labels.
To compensate for the incompleteness of the benchmark, we leverage the natural language inference (NLI) model to add correct but missing answers to the golden labels.
These two strategies ensure a comprehensive and adaptive evaluation, aligning closely with human judgment.

Through extensive experiments spanning three distinct information extraction tasks, we demonstrate the superiority of \ourmetric over baseline metrics, highlighting its efficacy in providing more human-aligned evaluations. 
Moreover, an in-depth analysis employing \ourmetric unveils nuanced insights into the performance of state-of-the-art LLMs, setting the stage for informed advancements in the field of information extraction.

\section{\ourmetric}
\subsection{Preliminaries and Overall Design}

\ourmetric aims to provide a fair evaluation for information extraction (IE) tasks, which extract structured knowledge from texts.
Formally, the input is a text $X = \{x_1, x_2, \ldots, x_n\}$, and the goal is to extract a set of information $Y = \{y_1, y_2, \ldots, y_m\}$ from $X$. 
We denote model predicted information as $\hat{Y}$.

The overall design of \ourmetric consists of two components, as shown in Figure~\ref{fig:pipeline}.
(1)~\textbf{Matcher $f(\cdot)$}. 
We fine-tune LLM to match every predicted information $\hat{y}_i$ with one and only one golden information $y_j$ if $\hat{y}_i$ is semantically consistent with $y_j$.
Namely, $f: \hat{Y} \mapsto Y$.
(2) \textbf{Complementer $g(\cdot)$}.
We leverage NLI models $g$ as complementers for alleviating the incompleteness of the dataset.
For each unmatched predicted information $\hat{y}_i$, we use NLI models to judge whether the given text $X$ entails the predicted information $\hat{y}_i$.
Specifically, $\hat{y}_i$ will be assigned an entailment score $s_i = g(X, \hat{y}_i)$.
If $s_i$ is greater than a threshold $\tau$, we consider the predicted information $\hat{y}_i$ should be added to the set of golden information $\hat{Y}$ with a weight $w_i = s_i\mathbbm{1}(s_i > \tau)$, where $\mathbbm{1}(\cdot)$ is the indicator function.

We calculate our modified precision and recall:
\small
\begin{align*}
    &\text{Precision} = \frac{1}{|\hat{Y}|}(\sum_{j=1}^n \mathbbm{1}(f(\hat{y}_i) = y_j) + \sum_{i=1}^m w_i) \\
    &\text{Recall} = \frac{1}{|Y|}(\sum_{j=1}^n \mathbbm{1}(f^{-1}(y_j) = \hat{y}_i) + \sum_{i=1}^m w_i)
\end{align*}
\normalsize
where $m$ is the number of unmatched predicted information and $n$ is the number of matched predicted information.
We define \ourmetric as the harmonic mean of this precision and recall.

\subsection{Matcher Construction}
\label{sec:llm_matcher}

We construct dataset $D = \left\{\left(Y, \hat{Y}, r\right)\right\}$ used to fine-tune LLM to behave as a matcher.
In particular, $r$ is the matching rationales for matching pairs $\left(\hat{y}_i, y_j\right)$.
It describes how the predicted information $\hat{y}_i$ is semantically equivalent to the golden information $y_j$.
We collect data from two different sources.

\textbf{Subjective Question Correction}.
Similar to IE tasks, the standard answer $Y$ of subjective questions from the Chinese College Entrance Examination (CCEE) also has multiple score points.
Teachers need to match from student answer $\hat{Y}$ to the score points to give the final score $s$.
This property makes the correcting subjective questions recorded by teachers in CCEE high quality for our task.

We collect triples $\{(Y, \hat{Y}, s)\}$ from mock tests of CCEE\footnote{We collect the data with grants from a high school.
These data points are processed, \textit{e.g.,} anonymized, to protect the privacy of students.
We will acknowledge the school by its name after the reviewing process due to the anonymization protocol of COLING-LREC.
}.
For correction data without the rationale to give the score, we prompt \texttt{gpt-3.5-turbo} to generate the rationale $r$ based on the collected triples $(Y, \hat{Y},s)$.
To ensure the data quality, we further filter data points to ensure:
(1) Each matching pair in $r$ must consist of one point from the student's answer and one point from the student's answer. 
(2) Every point in the student answer $\hat{Y}$ should only appear once in the matching pairs, so as the model answer $Y$.
(3) The number of matching pairs must be equal to the final score $s$.
We manually check the generated evaluation of 200 random filtered rationales and find that 98\% of them are correct, demonstrating the effectiveness of our strategies.

\textbf{Synthesised Data.}
We further enlarge the dataset with automatically synthesized data.
First, we build an answer point pool $P$ by assembling standard answer points from all the questions we collected.
Then, given one question $q$ and its corresponding model answer $Y$, we select $u$ points from the answer point pool $P$ as the negative points $\hat{Y}^{neg}$,
and $v$ points from the model answer $Y$ as the positive points $\hat{Y}^{pos}$.
Finally, we generate the student answer $\hat{Y} = \hat{Y}^{pos} \cup \hat{Y}^{neg}$.
To obtain highly-confusing negative points, we leverage BERTScore~\cite{bertscore} to select the most similar point but not in the model answer $Y$ as the negative points.
Finally, we use the template to generate the match rationales $r$.

Finally, we obtain $5,480$ data points from human subjective question correction and $4,460$ data points from synthesis. 
We train 7B variant of LLaMA-2~\cite{touvron2023llama2} and Tulu~\cite{tulu} for $2$ epochs.

\subsection{Complementer Construction}
To leverage NLI models as complementers, we use templates to generate the hypothesis $h_i$ for each unmatched predicted information $\hat{y}_i$.
For example, the template of Event Detection task is \texttt{This text describes a \{event triger\} event},
Then, we use the NLI model to compute the entailment score $s_i = g(X, h_i)$ for each predicted information $\hat{y}_i$ based on the given text $X$.
If the entailment score $s_i$ is greater than a threshold $\tau$, 
we believe the predicted information $\hat{y}_i$ should be added to the set of golden information $\hat{Y}$ with a weight $w_i = s_i\mathbbm{1}(s_i > \tau)$, where $\mathbbm{1}(\cdot)$ is the indicator function.
Specifically, to select one appropriate threshold $\tau$, we calculate the NLI-score of all the golden information $y$ from the training dataset of each task.
Then we select the $40$ percentile of the NLI-score of all the golden information $y$ as the threshold $\tau$.
We manually check the $200$ predicted information $\hat{y}_i$ whose entailment score $s_i$ is greater than the threshold $\tau$ and find that $80\%$ of them are correct. 
The correlation coefficient between entailment score and human judgment is $0.82$, which testifies to the plausibility of our strategies.

\section{Experiments}

We conduct experiments to validate the feasibility of \ourmetric.
Specifically, we evaluate several LLMs on different IE tasks and evaluate them via \ourmetric and baseline metrics.
We ask human annotators to select the most reasonable scores according to the input text, model output information, and golden information.

\noindent\textbf{Dataset:}\;\;
We evaluate \ourmetric on the task of Relation Extraction (RE) in NYT11~\cite{nyt11}, Event Detection (ED) and Event Argument Extraction (EAE) in ACE2005~\cite{ace2005}.

\noindent\textbf{Evaluated Models:}\;\;
We evaluate the following models.
$\bullet$~\textbf{Supervised Fine-tuned LLMs} includes 7B variant of Alpaca~\cite{alpaca}, Vicuna~\cite{vicuna} and Tulu~\cite{tulu}.
$\bullet$~\textbf{Reinforcement learning from human feedback (RLHF) LLMs} includes LLaMA-2-chat-7B~\cite{touvron2023llama2}.
Both kinds of LLMs are prompted with five examples following the format in chain-of-thought~\cite{wei2022chain}.
$\bullet$~\textbf{Task specific models} includes Text2Event~\cite{lu-etal-2021-text2event}---a sequence-to-sequence model combined with a constrained decoding algorithm to inject the event knowledge into the inference process---and UIE~\cite{lu-etal-2022-unified}---a unified model for IE tasks, which unifies different extraction structures via a structured extraction language and generates target extraction.

\noindent\textbf{Baseline Metrics:}\;\;
We compare \ourmetric with the following baseline metrics.
$\bullet$~\textbf{F1-score} is the most widely used evaluation metric in the IE area. It is defined as the harmonic mean of precision and recall.
$\bullet$~\textbf{BERTScore}~\cite{bertscore} is an embedding-based evaluation metric for text generation. It is defined as the cosine similarity between the contextual embeddings of the predicted text and the ground truth text.
$\bullet$~\textbf{BARTScore}~\cite{bartscore} also is an embedding-based evaluation metric for text generation.
It conceptualizes the evaluation of generated text as a text generation problem, modeled using pre-trained sequence-to-sequence models. 
It calculates the conditional probability of the output conditioned on the golden answer.
$\bullet$~\textbf{gpt-3.5-turbo} is prompted to evaluate the answer according to the input golden answer.

\noindent\textbf{Experiments Setup:}\;\;
For matchers in \ourmetric, we fine-tune and prompt Tulu and LLaMA-2 with one example.
For complementers in \ourmetric, the MENLI~\cite{chen_menli:2023} combined with MoverScore~\cite{zhao-etal-2019-moverscore} is employed.
For each task, we randomly select $800$ samples from all the predictions of the selected models as the corpora to evaluate \ourmetric.
We normalize all the metrics to the range of $[0, 1]$ for fair comparison.
Finally, for each sample, we ask human annotators to read the input text and the prediction.
Then, they are required to select (multiple) the best score from the scores of all the metrics.

\subsection{Experiment Results}

\begin{table}[t]
    \centering
    \scalebox{1}{
    \begin{tabular}{lccc}
    \toprule
    & {ED} & {RE} & {EAE} \\
    \midrule
    {BERTScore}  & $34.22$ & $21.37$ & $22.90$ \\
    {BARTScore}  & $19.44$ & $17.35$ & $22.14$ \\
    {GPT-3.5-Turbo}    & $45.20$ & $49.47$ & $51.40$ \\
    {F1-score}   & $50.00$ & $55.50$ & $62.72$ \\
    \midrule
    {\ourmetric\small{(Tulu)}}    & $54.80$ & $58.23$ & $67.05$ \\
    {\ourmetric\small{(LLaMA-2)}} & $57.07$ & $61.39$ & $67.05$ \\
    \bottomrule
    \end{tabular}
    }
    \caption{
    Human preference rate (\%) of metrics.
    }
    \label{tab:main-result}
\end{table}

We show the results in Table \ref{tab:main-result}.
The results show that \ourmetric is more preferred by human annotators than the baseline metrics.

Specifically, \ourmetric with LLaMA-2 (fine-tuned) achieves the best performance on all the tasks. Then, \ourmetric with Tulu (fine-tuned) achieves the second-best performance on all the tasks.
Besides the metrics designed for text generation (\textit{i.e.,} BERTScore, BARTScore) do not perform promisingly on the IE tasks, even compared to the F1-score.
This validates our initial hypothesis that these metrics designed for text generation are not suitable for IE tasks.

\subsection{Revisiting LLMs for IE}

With the help of \ourmetric, we revisit the performance of existing state-of-the-art conventional models and LLMs on IE tasks.

The results are shown in Table~\ref{tab:revisit-performance}.
In general, compared to the F1-score, \ourmetric is relatively higher for both LLMs and conventional models.
Specifically, in the ED and RE tasks, for the performance of LLMs, the score of \ourmetric is higher than the F1-score by $+20\%$ and $+30\%$ on average respectively.
It indicates that the F1-score underestimates LLMs to some degree.
This big increase brings the performance of LLMs closer to conventional models.
Considering that there is no task-specific fine-tuning for LLMs, this result is promising, which shows the potential of LLMs in some shallow IE tasks.
As for the EAE task, the increase of \ourmetric is relatively small about $+10\%$.
Besides, the performance of LLMs is still far behind the conventional models.
Compared to the former two tasks, EAE task requires LLMs to have a deeper understanding of eventsp, specifically, the event schema.
This discrepancy indicates that LLMs may still struggle to strictly extract information with a well-defined schema.

\begin{table}[t]
    \renewcommand{\arraystretch}{1.2}
    \scalebox{0.83}{
    \setlength{\tabcolsep}{2.5pt}
    \begin{tabular}{lccccc}
    \toprule
    MODEL                         & ED & RE & EAE \\ 
    \midrule
    Alpaca              
       & \cellcolor{mgreen} $27.3$ ($+12.0$)  & \cellcolor{ddgreen} $46.9$ ($+38.2$)    & \cellcolor{sgreen}$\;\;6.2$ ($+\;\;5.5$)    \\
    Vicuna           
       & \cellcolor{dgreen} $36.2$ ($+29.2$)  & \cellcolor{ddgreen} $37.4$ ($+34.5$)    & \cellcolor{sgreen} $10.3$ ($+\;\;9.8$)    \\
    Tulu            
       & \cellcolor{dgreen} $37.0$ ($+29.3$)   & \cellcolor{ddgreen} $40.6$ ($+35.7$)    & \cellcolor{sgreen} $\;\;9.1$ ($+\;\;8.8$)       \\
    LLaMA-2-Chat             
       & \cellcolor{dgreen} $36.9$ ($+28.3$)  & \cellcolor{ddgreen} $42.3$ ($+38.2$)     & \cellcolor{mgreen} $12.1$ ($+11.8$)   \\         
    \midrule  
    SOTA
       & \cellcolor{sgreen} $77.5$ ($+\;\;9.1$)  & \cellcolor{sgreen} $89.4$ ($+\;\;2.7$)     & \cellcolor{dgreen} $71.0$ ($+19.9$)   \\ 
    \bottomrule
    \end{tabular}
    }
    \caption{
    Model performance on the ED, RE, and EAE datasets.
    The numbers in parenthesis represent the difference between \ourmetric and the F1-score.
    The selected SOTA model is Text2Event for ED and EAE, and UIE for RE.
    }
    \label{tab:revisit-performance}
\end{table}

\section{Related Works}

Traditional evaluation metrics in IE tasks include Precision, Recall, and F1-score, which are foundational in the assessment of IE systems.
However, these metrics are static, struggling to handle semantic-level consistency between the extracted information and the ground truth.
Recently, some dynamic metrics from natural language generation such as BLEU \cite{bleu}, ROUGE \cite{rouge}, BERTScore \cite{bertscore}, and BARTScore \cite{bartscore} are introduced into the evaluation of IE tasks.
However, these metrics are not suitable, 
as they essentially measure the similarity between two texts, ignoring the precision and completeness requirements in IE tasks.

\section{Conclusion}
In this paper, we propose a novel evaluation method, \ourmetric.
It leverages LLMs as matches and NLI models as complementers to handle imprecision and incompleteness in the conventional evaluation methods.
Through human evaluation, we testify that \ourmetric is more preferred by human annotators than the baseline methods.
With \ourmetric, we re-evaluate the performance of state-of-the-art LLMs on IE tasks and provide insights for future research that (1) LLMs are capable of conducting shallow IE tasks, and (2) LLMs still struggle to structurally extract information with well-defined schema.

\section*{Code and Data Availability Statement}
The artifacts associated with this paper include both datasets and experiment codes.
 
For dataset construction, it is preferred to ask human annotators to give each subjective question correction a detailed rationale explaining how this question is scored. 
However, considering the high cost and extensive labor work, we use the \texttt{gpt-3.5-turbo} to obtain the rationale based on some filtering rules. 
Also, we enrich the original dataset by combining initial gold answer with the most confusing answer point within the answer point pool. 
By further human checking, we testify the effectiveness and truthfulness of our dataset. 
Thus, the full dataset includes (1) the subjective question correction dataset and (2) the synthesis dataset.

Our codes include scripts that are used to generate data, and codebases that are used to establish the initial baselines. 

Per the request of anonymous protocol, we will release both the subjective question correction dataset, the synthesis dataset and checkpoints of fine-tuned LLMs as soon as this paper is accepted after the reviewing process.

\section*{Ethical Consideration}
We primarily focus on two ethical issues: the privacy of collected data and the treatment of annotators

For the privacy of collected data, 
part of data is collected from a high school with grant.
These data points are processed, \textit{e.g.,} anonymized, to protect the privacy of students.
We will acknowledge the school by its name after the reviewing process due to the anonymization protocol of COLING-LREC.

For the treatment of annotators, they are all students undertaking post-secondary educations.
The annotator work is conducted in their part-time.
We provide hourly wages that meet the local average for data work, ensuring equal pay for equal work across all employees.

\section*{Limitations}
For LLMs as Matchers part in \ourmetric, 
we fine-tune the LLMs on the subjective question correction dataset.
Subsequently, they are employed to match predicted information with the golden information in IE task.
But there is a discrepancy between subjective questions from CCEE and data in IE task.
Particularly, although the answer of them are both consisted of a set of key points, 
while the point in subjective questions is a single sentence, while in IE task, the point is usually a phrase or a word.
Thus the matching in subjective question correction is between sentences, which is more difficult than matching between phrases or words in IE task.
But we empirically find that the fine-tuned LLMs still work well on matching the predicted information with the golden information in IE task.
This is because training on more challenging tasks (sentence matching) has endowed large language models with robust generalization performance, enabling them to excel in relatively simpler tasks (phrase or word matching).

\section*{Acknowledgement}

This work is supported by a grant from the Institute for Guo Qiang, Tsinghua University (2019GQB0003), Zhipu AI, and the Tsinghua University (Department of Computer Science and Technology)-Siemens Ltd., China Joint Research Center for Industrial Intelligence and Internet of Things (JCIIOT).

Additionally, we would like to express our gratitude to Taixing High School of Jiangsu Province for sharing their invaluable examination data.
It would be impossible to construct our dataset without their supportance.

\clearpage

\section*{Bibliographical References}
\bibliography{1-ref}
\bibliographystyle{lrec_natbib}

\clearpage

\appendix
\section{Dataset Construction}

Typically, every record in our dataset consists of ``reference answer'',   ``student answer'', ``total score'', and ``grading process'' four parts.
To obtain the grading process, we manually write several grading processes and use In-Context Learning to help GPT-3.5-turbo generate.

\begin{table}[ht]
    \centering
    \scalebox{0.72}{
    \begin{tabular}{p{1.3\linewidth}}
        \toprule
        \vspace{-2mm}
        \textbf{\textsc{Instruction:}} \\
            Assuming you're a teacher, please describe the grading process based on a student's response. \\
            I will present a ``standard answer'', a ``student response'', a ``total points for the question'', and a ``score obtained by the student''. \\
            You need to explain how the student's score was derived. Note that your task is not to grade the example, but to explain how the score was obtained. \\
            Start the explanation with ``Grading Process''. In each step, include the content from the text and the answer, as well as the score. \\
        \bottomrule
    \end{tabular}
}
        \caption{
  The instruction for generating the grading process.
    }
    \label{tab:grading_process}
\end{table}

The grading process is in the form of ``\textit{s} in the reference answer corresponds to \textit{t} in the student answer and earns \textit{k} points.'' where \textit{s} is a point in the reference answer, \textit{t} is a point in the student answer, and \textit{k} is the mark when you get to the point.

\section{Ablation Study}

\subsection{Performance on Different Datasets}

Our CCEE dataset consists of two parts, the natural student answer and the synthesized student answer. To prove the effectiveness of the synthesized part, we fine-tuned our evaluation model on the natural parts and on the mixed parts, \textit{i.e.,} mixing the natural parts and the synthesized parts, and then evaluate them based on the same dev dataset.

\begin{table}[ht]
    \centering
    \scalebox{0.72}{
    \begin{tabular}{lccc}
        \toprule
        & untuned & natural-data-tuned & full-data-tuned \\
        \midrule
    Vicuna-7b & 7.5\% & 14.0\% & 46.2\% \\
        \midrule
    Llama2-7b-chat & 11.0\% & 36.6\% & 41\% \\
        \midrule
    Tulu-7b & 7\% & 13.1\% & 42.9\% \\
        \bottomrule
    \end{tabular}
    }
    \caption{Results on dev dataset. The percentage value is the accuracy rate where the model gives exactly the same scores as the teacher did.}
    \label{results:fine-tuned}
\end{table}

The synthesized dataset improves the performance of evaluation models by a large margin, proving its effectiveness.

\subsection{NLI Effectiveness}

To show the results of LLM-Matcher and NLI-Complementer separately, we conduct an ablation study.
\begin{table}[htbp]
\scalebox{0.66}{
\centering
\begin{tabular}{c|cc|cc|cc}
\toprule
& \multicolumn{2}{c|}{F1-Score} & \multicolumn{2}{c|}{SQC-Score without NLI} & \multicolumn{2}{c}{SQC-Score} \\
\small
& RE & ED & RE & ED & RE & ED \\
\midrule
Alpaca-7b & 15.4 & 8.3 & 26.9 & 22.2 & 27.3 & 46.9 \\
\midrule
Vicuna-7b & 7.0 & 3.0 & 26.6 & 19.6 & 36.2 & 37.4 \\
\midrule
Tulu-7b & 7.7 & 4.3 & 27.6 & 23.6 & 37.0 & 40.6 \\
\midrule
Llama2-chat-7b & 8.6 & 3.8 & 28.2 & 22.3 & 36.9 & 42.3 \\ 
\bottomrule
\end{tabular}
}
\caption{
Scores of different evaluation metrics on RE and ED. We show the results of F1-Score, SQC-Score without NLI complemeneter and full SQC-Score. 
}
\label{ablation study}
\end{table}

Using SQC-Score, compared with traditional F1-Score, results in a better performance which corresponds to our intuition. Also, with NLI to remedy the annotation of datasets, the performance will be better improved.

\section{NLI Justification}

We assume that manual labels are of great value, which implies that they should be entailed by the original text with a high NLI Score. 

\begin{table}[ht]
    \centering
    \scalebox{1.0}{
    \begin{tabular}{lcccccc}
        \toprule
        & 55\% & 50\% & 45\% & 40\% & 35\% & 30\% \\
        \midrule
    ED & 86.0 & 87.5 & 88.8 & 90.1 & 91.3 & 92.5 \\
        \midrule
    RE & 80.1 & 82.3 & 84.6 & 86.8 & 88.7 & 90.4 \\
        \midrule
    EAE & 49.0 & 51.7 & 54.6 & 57.8 & 61.4 & 65.2  \\
        \bottomrule
    \end{tabular}
    }
    \caption{
    NLI results of topk\% gold labels
    }\label{NLI}
\end{table}

We propose to use NLI to remedy the drawbacks of the annotation of IE datasets. To make it specific, we choose the NLI score of the top 40\% gold labels as a threshold. Any prediction whose NLI score is above that is considered reasonable but not annotated. \\

\newpage
\section{Case Study}

In our training data, we treat the gold as reference answers and the model predictions as student answers. By assigning some same points to each triple, and giving the whole scores, we can prompt LLMs to give the final scores the ``student answers``. We transfer this type of evaluation to IE tasks.

\begin{table}[ht]
    \centering
    \scalebox{0.72}{
    \begin{tabular}{p{1.3\linewidth}}
        \toprule
        \vspace{-2mm}
        \textbf{\textsc{Instruction:}} \\
        Assuming you are a teacher, please evaluate the student's answer and provide the corresponding score. \\
        Now, I will show you a "Standard Answer", a "Student Answer", and the "Total Score" for the question. This is all the information you need to grade the student's answer. \\
        You need to tell me how many points this student's response received and how it was derived.  \\
        Note, that you're not only scoring the example, but you also need to explain how the score was determined, detailing every step of the evaluation process by providing the content from the student's answer and the standard answer, as well as the score.  \\
        The student score should not exceed the standard score. \\
        \midrule
               \textbf{\textsc{Input:}} \\
        Standard Answer:\\
        Attack Trigger War;(2 points) \\
        Attack Attacker Tom;(2 points) \\ 
        Attack Method Bomb(2 points)\\
        Student Answer: \\
        Attack Person Tom; \\
        Transport Trigger War; \\
        Attack Result Explosion.\\
        Total Score: \\
        6 points \\
        \midrule
               \textbf{\textsc{Output:}}  \\
        Grading Process:\\
        Let's think step by step ``attack Attacker Tom`` in the standard answer corresponds to the ``Attack Person Tom`` in the student answer, earning 2 points.\\
        ``Attack Result Explosion`` in the student answer corresponds to the ``Attack Method Bomb`` in the standard answer, earning 2 points.\\
        There is no information in the student's answer that matches ``Attack Trigger War`` and no points are given.\\
        Therefore, the final score is 4 points.\\
        \bottomrule
    \end{tabular}
    }
        \caption{
  The instruction and an example of SQC-Score
    }
    \label{tab:exmaple}
\end{table}

\newpage
\section{Instructions for Information Extraction}

\subsection{Relation Extraction}

\begin{table}[ht]
    \centering
    \scalebox{0.72}{
    \begin{tabular}{p{1.3\linewidth}}
        \toprule
        \vspace{-2mm}
        \textbf{\textsc{Instruction:}} \\
        Task: Identify the relationships and their corresponding entities within the provided sentence. \\
        Note: \\
        - The relation should be one-word-long or two-word-long. \\
        - Respond in several tuples: (head entity 1, relation 1, tail entity 1), (head entity 2, relation 2, tail entity 2)... \\
        - If no relations or entities are identified, answer: (none, none, none). \\
        \bottomrule
    \end{tabular}
}
        \caption{
  The instruction for Relation Extraction.
    }
    \label{tab:re1}
\end{table}

\subsection{Event Detection}

\begin{table}[ht]
    \centering
    \scalebox{0.72}{
    \begin{tabular}{p{1.3\linewidth}}
        \toprule
        \vspace{-2mm}
        \textbf{\textsc{Instruction:}} \\
            Your task is to extract events from the given text. \\
            An event is a specific occurrence involving participants.\\
            If no events in the given text, respond: none. \\
            Note: \\
            The event type should be an exactly one-word noun or an exactly one-word verb.\\
            Respond in a tuple format, e.g. (1. specific event type 1, 2. specific event type 2, ......).\\
        \bottomrule
    \end{tabular}
}
        \caption{
  The instruction for Event Detection.
    }
    \label{tab:ed}
\end{table}

\subsection{Event Argument Extraction}

\begin{table}[ht]
    \centering
    \scalebox{0.72}{
    \begin{tabular}{p{1.3\linewidth}}
        \toprule
        \vspace{-2mm}
        \textbf{\textsc{Instruction:}} \\
Please analyze the provided text enclosed between "||" symbols and systematically extract relevant event arguments based on corresponding roles. Some roles are listed below: \\
- Event Type: The primary category or nature of the event. \\
- Trigger: The word or phrase that indicates the occurrence of the event. \\
- Time: When the event took place. \\
- Person: Any individual or entity involved in the event. \\
- Location: Where the event took place. \\
- Action: What was done during the event? \\

Each extracted detail should be represented as a tuple in the format:\\
(specific event type, specific argument role, specific content)\\

In cases where a specific role does not have identifiable content in the text, use 'none' as the placeholder.\\

Your answer should be organized in a clear, structured manner, following the sequence of the extracted details. Please ensure that each tuple is separated by a comma.\\
        \bottomrule
    \end{tabular}
}
        \caption{
  The instruction for Event Detection.
    }
    \label{tab:re2}
\end{table}

\end{document}